\documentclass[sigconf]{acmart}
\AtBeginDocument{%
  \providecommand\BibTeX{{%
    \normalfont B\kern-0.5em{\scshape i\kern-0.25em b}\kern-0.8em\TeX}}}

\copyrightyear{2023}
\acmYear{2023}
\setcopyright{acmlicensed}
\acmConference[MM '23] {Proceedings of the 31st ACM International Conference on Multimedia}{October 29--November 3, 2023}{Ottawa, ON, Canada.}
\acmBooktitle{Proceedings of the 31st ACM International Conference on Multimedia (MM '23), October 29--November 3, 2023, Ottawa, ON, Canada}
\acmPrice{15.00}
\acmISBN{979-8-4007-0108-5/23/10}
\acmDOI{10.1145/3581783.3612511}
\usepackage{bm}
\usepackage{multirow}
\usepackage{multicol}
\usepackage{enumitem}
\usepackage{makecell}
\usepackage{balance}

\begin{document}

\title{Class-level Structural Relation Modelling and Smoothing for Visual Representation Learning}


% \author{Zitan Chen}
% \affiliation{%
%   \institution{Shandong University}}
% \email{chenzt@mail.sdu.edu.cn}

% \author{Zhuang Qi}
% \affiliation{%
%   \institution{Shandong University}}
% \email{z_qi@mail.sdu.edu.cn}

% \author{Xiao Cao}
% \affiliation{%
%   \institution{National University of Singapore}}
% \email{xiaocao@u.nus.edu}

% \author{Xiangxian Li}
% \affiliation{%
%   \institution{Shandong University}}
% \email{xiangxian_lee@mail.sdu.edu.cn}

% \author{Xiangxu Meng}
% \affiliation{%
%   \institution{Shandong University}}
% \email{mxx@sdu.edu.cn}

% \author{Lei Meng}
% \authornote{Corresponding author.}
% \affiliation{%
%   \institution{$^1$Shandong University}
%   \institution{$^2$Shandong Research Institute of Industrial Technology}}
% \email{lmeng@sdu.edu.cn}

\author{Zitan Chen}
\email{chenzt@mail.sdu.edu.cn}
\affiliation{%
  \institution{Shandong University}
  \country{}}

\author{Zhuang Qi}
\email{z_qi@mail.sdu.edu.cn}
\affiliation{%
  \institution{Shandong University}
  \country{}}

\author{Xiao Cao}
\email{xiaocao@u.nus.edu}
\affiliation{%
  \institution{National University of Singapore}
  \country{}}

\author{Xiangxian Li}
\email{xiangxian_lee@mail.sdu.edu.cn}
\affiliation{%
  \institution{Shandong University}
  \country{}}

\author{Xiangxu Meng}
\email{mxx@sdu.edu.cn}
\affiliation{%
  \institution{Shandong University}
  \country{}}

\author{Lei Meng}
\email{lmeng@sdu.edu.cn}
\authornote{Corresponding author.}
\affiliation{%
  \institution{$^1$Shandong University}
  \institution{$^2$Shandong Research Institute of Industrial Technology}
  \country{}}

\renewcommand{\shortauthors}{Zitan Chen et al.}

\begin{abstract}
 Representation learning for images has been advanced by recent progress in more complex neural models such as the Vision Transformers and new learning theories such as the structural causal models. However, these models mainly rely on the classification loss to implicitly regularize the class-level data distributions, and they may face difficulties when handling classes with diverse visual patterns. We argue that the incorporation of the structural information between data samples may improve this situation. To achieve this goal, this paper presents a framework termed \textbf{C}lass-level \textbf{S}tructural \textbf{R}elation \textbf{M}odeling and \textbf{S}moothing for Visual Representation Learning (CSRMS), which includes the Class-level Relation Modelling, Class-aware Graph Sampling, and Relational Graph-Guided
Representation Learning modules to model a relational graph of the entire dataset and perform class-aware smoothing and regularization operations to alleviate the issue of intra-class visual diversity and inter-class similarity. Specifically, the Class-level Relation Modelling module uses a clustering algorithm to learn the data distributions in the feature space and identify three types of class-level sample relations for the training set; Class-aware Graph Sampling module extends typical training batch construction process with three strategies to sample dataset-level sub-graphs; and Relational Graph-Guided Representation Learning module employs a graph convolution network with knowledge-guided smoothing operations to ease the projection from different visual patterns to the same class. Experiments demonstrate the effectiveness of structured knowledge modelling for enhanced representation learning and show that CSRMS can be incorporated with any state-of-the-art visual representation learning models for performance gains. The source codes and demos have been released at \url{https://github.com/czt117/CSRMS}.
\end{abstract}







\maketitle

\section{Introduction}
\begin{figure}[t]
\centerline{\includegraphics[width=1.0\linewidth]{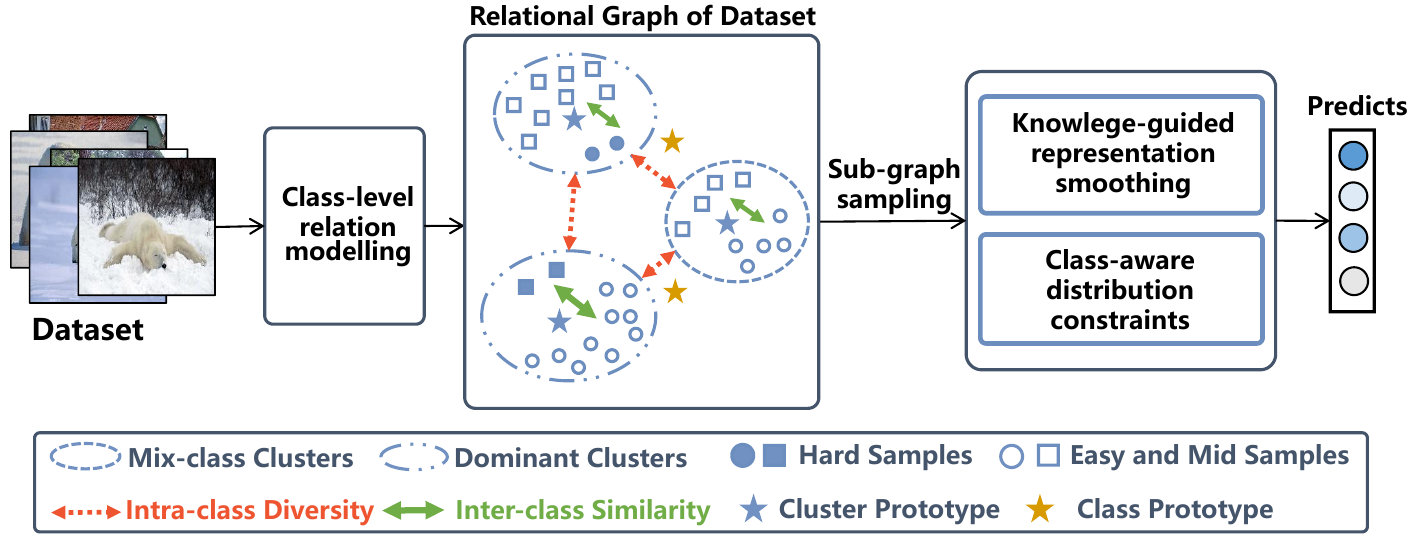}}
\vspace{-0.3cm}
\caption{Illustration to CSRMS. CSRMS constructs a dataset-level relation graph using visual patterns, conducts sub-graph sampling based on relations, and performs representation learning with knowledge smoothing and class-level constraints to aggregate information.}
\label{fmotivation2}
\vspace{-0.75cm}
\end{figure}
% Image classification is an important basis in computer vision tasks like object detection \cite{c1,c2}, face recognition \cite{c3,c4}, and pose estimation \cite{c5,c6}. 
In recent years, deep learning has witnessed remarkable advancements across various fields, including classification\cite{cls1,cls2,cl3,cls4,cls5,cls6,cls7}, recommendation\cite{RS1, RS2, RS3, RS4, RS5}, image generation\cite{gen2,gen3,gen4,gen6} federal learning\cite{fl1,fl2} and video analysis\cite{yang1,yang2,yang3,yang4,yang5}.
Currently, deep learning-based image classification methods typically involve extracting visual representations using a visual backbone network and then mapping these representations to their corresponding classes. However, the high complexity of visual information poses a challenge for the model to extract discriminative representations for each class effectively. Previous works have primarily focused on increasing the complexity of the network architecture at the model level \cite{c7,c8,c9} or introducing instance-level image transformations to enhance the model's learning capacity \cite{c10,c11,c12,c13}. Although these approaches have brought incremental improvements, the intra-class diversity problem in the image limits the models' learning of representations in each class, and the inter-class similarity also affects the decision-making of models.

To mitigate the aforementioned issues, it would be beneficial to clarify the role of samples in model training as well as the interrelation among samples of each class. This can assist the model in achieving more effective optimization. However, the exploration of sample relationships is still in its early stages. Existing methods mainly focus on exploring instance-level relations between samples, such as identifying outlier representations through clustering algorithms and performing multiple rounds of iteration for correction \cite{c40}, or using attention mechanisms or graph construction methods to implicitly explore the relations between samples and enhance representation learning \cite{c14,c15,c18,c19}. Although these methods can utilize intra-class information to improve intra-class diversity, the attention to inter-class similarity is still inadequate. Contrastive learning \cite{c22,c23} is a method that explores the relationships between images by using images from the same class as positive samples and images from different classes as negative samples, to narrow the gap between images from the same class and increase the gap between images from different classes. However, how to delve deeper into the relationships and construct reasonable and efficient constraint methods is still a challenge. Existing methods often exhibit inadequate sample selection and do not pay sufficient attention to the difficulty of representation learning, which limits their ability to handle complex image data.

\begin{figure}[t]
\centerline{\includegraphics[width=1.0\linewidth]{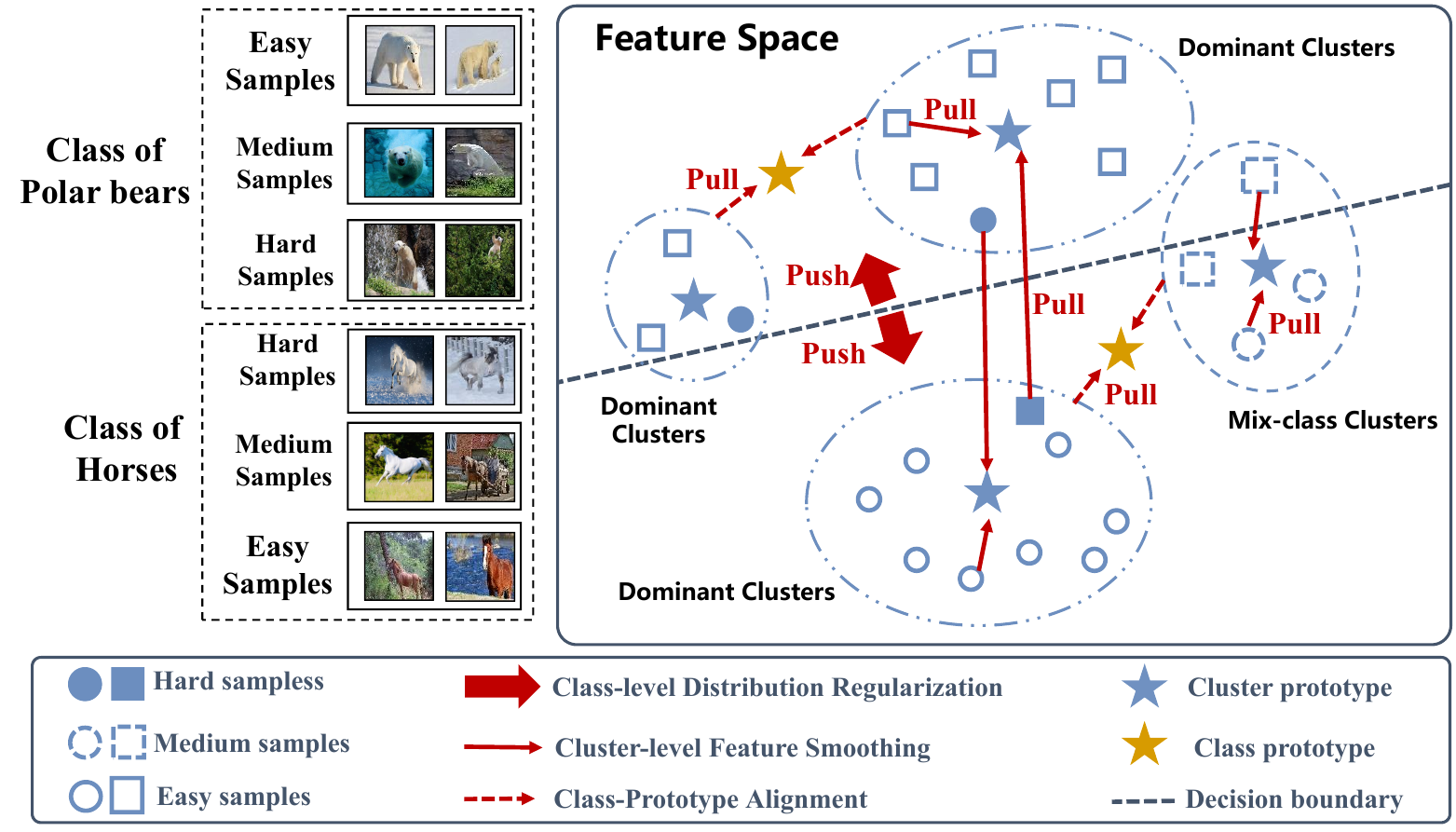}}
\vspace{-0.4cm}
\caption{
Illustration to strategies of CSRMS to mitigate intra-class diversity of visual patterns.}
\label{fmotivation}
\vspace{-0.65cm}
\end{figure}

To address these limitations, we propose a novel approach for enhancing image representation, named CSRMS, to address the aforementioned challenges. The main concept of the approach is presented in Figure \ref{fmotivation2}. The method constructs dataset-level hierarchical relation graphs by mining class-level knowledge, thereby achieving an explicit and effective sample relationship. In the Offline Learning Process (OLP) stage, CSRMS explores the distribution patterns of images and constructs the relation graph to guide the representation learning. In the Training Process (TP) stage, Class-aware Graph Sampling(CaGS) is conducted to acquire a batch-level sub-graph from the dataset-level hierarchical graph, based on sampling strategies. Moreover, curriculum construction is conducted based on easy-hard estimation to further enhance the representation learning. Finally, in Relational Graph-Guided Representation Learning (RGRL) the above relations are constrained and regularized visual representations are generated through graphical smoothing and class-level constraint. 
% The relationship between the visual representations of images at the class-level prevents the construction of sample relationships in a "regional" manner, avoiding the neglect of the association between some samples. 
By constructing sampling strategies, "selective" sampling can be performed, reducing redundancy and enhancing the efficiency of representation learning. Through the construction of curriculum and relational graph-guided representation learning, representations can be regularized progressively, which can alleviate the problems of intra-class differences and inter-class similarities and achieve better representation learning effects as illustrated in Figure \ref{fmotivation}.

Experiments are conducted on the CIFAR10, CIFAR100, Vireo172, and NUS-WIDE datasets in terms of performance comparison. Ablation study of the key components of CSRMS and in-depth analyses of different positive and negative sampling strategies demonstrate that CaGS and RGRL can better model sample relations, construct constraints of representations and then continuously improve classification accuracy. Case studies for visual distribution patterns and behaviours in different successful and failure cases of CSRMS demonstrate that CSRMS can effectively alleviate the problems of intra-class differences and inter-class similarities. To summarize, the main contributions are:
% \vspace{-0.4cm}
\begin{itemize}[leftmargin=10pt]
\item  A model-agnostic framework CSRMS is proposed, which models a relational graph and performs class-aware smoothing and regularization to address inter-class similarity and intra-class diversity, resulting in enhanced classification accuracy.
\item A novel class-level strategy is proposed, which models the class-level relationships of visual representations, constructs effective sampling strategies and curriculum and utilizes explicit smoothing and constraints to enhance representation learning.
\item Experiments demonstrate that the relation-based sampling and smoothing, the class-aware regularization can effectively alleviate intra-class diversity and inter-class similarity, and improve classification performance. This verifies the effectiveness of CSRMS in enhancing representation learning.
\end{itemize}

\section{Related Work}
\subsubsection*{Sample Relationship in Image Classification.}
The relations between samples have been extensively studied using implicit or explicit approaches. The former \cite{c14,c15,c16} leverages cross-attention mechanisms, such as transformers or attention modules, to implicitly capture batch-level inter-sample relationships. Explicit schemes, such as Graph Neural Networks (GNN), explicitly model the relation graph among same-class samples to enhance visual representations \cite{c18,c19,c20}. However, these methods only address intra-class diversity, while neglecting inter-class similarity. Contrastive learning methods \cite{c22,c23,c24} explicitly explore sample relationships through the creation of a unified contrast constraint, pushing apart representations of different categories while drawing closer those of the same class. Despite the effectiveness of contrastive learning in mitigating both intra-class diversity and inter-class similarity, current approaches generally lack efficient sampling strategies for constructing comparison constraints, which may limit their efficiency and capability for representation learning.

\begin{figure*}[t]
\centerline{\includegraphics[width=1.0\linewidth]{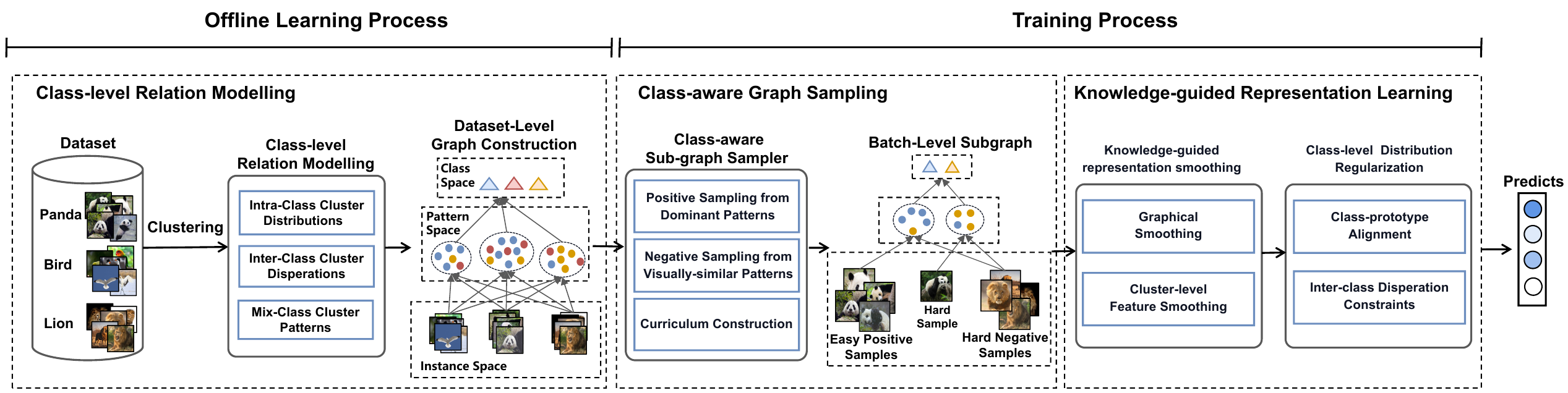}}
\vspace{-0.2cm}
\caption{
Illustration of CSRMS. CSRMS constructs a dataset-level relation graph by exploring the class-level relations. The Class-aware Graph Sampling module constructs a relational sub-graph based on class-aware curriculum sampling. The Relational Graph-Guided Representation Learning module acquires enhanced representations by smoothing and regularization.}
\label{frame}
\vspace{-0.3cm}
\end{figure*}

\subsubsection*{Class-aware Information Guided Image Classification.}
The training of classification models in conventional methods heavily relies on classification loss, which can be easily disrupted by differences within classes and similarities between classes. To address this problem, class-aware methods extract extra class-level information for enhancing image classification and approaches can be typically divided into two categories. The methods of the first type utilize the classification information generated by the model to integrate class-specific representations, such as prototypes, into representation learning \cite{DBLP:conf/miccai/YangPYSZZB22,DBLP:conf/cvpr/ZhuWCCJ22,cheng2023class}, or as a constraint for improving contrastive learning optimization \cite{caron2020unsupervised}. The methods of the second type utilize class labels as guidance, such as clustering to obtain the visual-spatial distribution of images, followed by introducing label information to constrain model learning for specific clusters \cite{DBLP:conf/eccv/XuLLL22}; other methods in this approach involve incorporating labels into the constraints of contrastive learning \cite{DBLP:conf/aaai/ZhaoZWS022}, augmenting images regarding the class \cite{c39}, or using labels as more detailed guiding conditions in multi-label classification \cite{DBLP:conf/icme/ChenWJG19,liu2023incomplete}.

% \vspace{-1.5cm}
\section{PROBLEM FORMULATION}
In conventional image classification algorithms, the dataset comprises a set of images $\mathcal{I}=\{I_i|i=1,2...N\}$ and corresponding labels $\mathcal{Y}=\{y_i|i=1,2...N\}$. Traditional deep learning-based methods usually learn a visual mapping $\mathcal{M}_{v}(\cdot)$ for visual inference $I_{i} \rightarrow \textbf{F}_{i}$, where $\textbf{F}_{i}$ represents the visual representation vector of image $I_{i}$. Subsequently, conventional algorithms generally train a classifier $\mathcal{C}(\cdot)$ for predicting the class of $I_{i}$, i.e., $\textbf{F}_{i} \rightarrow \textbf{P}_{i}$, where $\textbf{P}_{i}$ denotes the predicted scores associated with image $I_{i}$. During the training, a classification loss function, such as cross-entropy $\mathcal{L}_{ce}$, is employed to limit the discrepancy between the prediction and the label.

Different from conventional approaches, the proposed CSRMS firstly clusters visual representations $\textbf{F}$ and incorporates class information $\mathcal{Y}$ to create a relation graph $G_{t}$ of the training set offline. During the online training, CSRMS extends the typical batch construction process to sample sub-graphs $G_{b}$ from $G_{t}$.
By leveraging a graphical convolutional mapping $\mathcal{G}(\cdot,\cdot)$, CSRMS performs knowledge-guided graphical smoothing to form better representations $\textbf{F}_{G}$, i.e., $\mathcal{G}(G_{b}, \textbf{F}) \rightarrow \textbf{F}_{G}$. To achieve smooth representations, CSRMS introduces cluster prototypes $w_{cluster}$ and class prototypes $w_{class}$ for cluster-level alignment $\textbf{F}_{G} \odot w_{cu} \rightarrow \textbf{F}_{u}$ and class-level alignment $\textbf{F}_{u} \odot p_{ca} \rightarrow \textbf{F}_{a}$. Finally, CSRMS trains a classifier $\mathcal{C}(\cdot)$ for class predictions $\textbf{F}_{a} \rightarrow \textbf{P}$, where $\textbf{P}$ represents the predicted scores of images.

\section{METHOD}
The CSRMS proposes a targeted approach to alleviate the issue of intra-class visual diversity and inter-class similarity. The pipeline of CSRMS is illustrated in Figure \ref{frame}.
Specifically, during the Offline Learning Process, the Class-level Relation Modelling module uses a clustering algorithm to learn the data distributions in the feature space and identify three types of class-level sample relations for the training set. In the Training Process, the Class-aware Graph Sampling module extends the typical training batch construction process with three strategies to sample sub-graphs; and the Relational Graph-Guided Representation Learning module employs a graph convolution network with knowledge-guided smoothing operations to ease the projection from different visual patterns to the same class. 

\subsection{Class-level Relation Modeling (CRM)}
To model the relationships among samples, CSRMS utilizes adaptive clustering in the CRM module to identify the aggregation of samples in the visual space, thereby capturing the distribution of patterns in the dataset. CRM then incorporates the class information of the samples to extract the "class-pattern-instance" relationships and organize them into a relational graph.

\subsubsection{Definations of Class-Level Sample Relations}\label{411}
To clarify the relationship between samples and further strengthen their coherence, we first identify three types of bad relationships based on their feature distributions during clustering and their labels. For samples $I_a$ and $I_b$, the relationship is defined as follows:

\begin{enumerate}[leftmargin=10pt]
    \item [(a)] \textbf{Intra-Class Diversity of Visual Patterns}: Samples within a class usually scatter across different dominant clusters, i.e., clusters that dominantly consist of samples in a certain class, in the visual space. These hard samples make it challenging to learn the class's representation and this relation is defined as:
    \begin{equation}
        R_{ID}=\{(I_{a},I_{b})|y_{a}=y_{b},PC_{a}\neq PC_{b}\}
    \end{equation}
    where $PC_a$ and $PC_b$ represent the dominant clusters that $I_{a}$ and $I_{b}$ belonging to, respectively.
    \item [(b)] \textbf{Inter-Class Similarity of Visual Patterns}: In dominant clusters, visually similar samples belonging to different classes can make it difficult to differentiate. That can be defined as:
    \begin{equation}
        R_{IS}=\{(I_{a},I_{b})|y_{a} \neq y_{b},PC_{a}=PC_{b}\}
    \end{equation}
    \item [(c)] \textbf{Mix-class Cluster of Visual Patterns}: Since the variability in vision, visually-clustered clusters may not have a dominant class. Such mixed-class clusters are complex to model, as they involve the relationship between different sample classes. This relationship can be termed as:
    \begin{equation}
        R_{MC}=\{(I_{a},I_{b})|y_{a} \neq y_{b},MC_{a}=MC_{b}\}
    \end{equation}
    where $MC_a$ and $MC_b$ represent the mix-class clusters that $I_{a}$ and $I_{b}$ belonging to, respectively.
\end{enumerate}

\subsubsection{Relational Graph Construction of Dataset.}
To utilize and enhance the relationship outlined in Section 4.1.1, we depict the samples on a dataset-level relational graph $G_{t}$, which comprises class level, pattern level, and instance level.
The connections between instance level and class level of $G_{t}$ rely on the corresponding between images $I$ and their labels $Y$. At the pattern level, the adaptive clustering algorithm ART \cite{c41} constructs the feature space and identifies clusters C to represent patterns based on the aggregation relationship of samples.
\begin{equation}
    C = \operatorname{ART}(\mathcal{M}_{v}(I))
\end{equation}
where $\mathcal{M}_{v}(\cdot)$ is the visual mapping, the clusters $C=\{c_1,...,c_J\}$ and $J$ is the number of clusters.
Consequently, $G_t$ can be represented by $\{Y, C, I\}$. For each instance $I_i$ that is labeled $y_i$, we can get the pattern $c_j$ by matching the clusters and find the connection $I_{i}\rightarrow c_{j}\rightarrow y_{i}$.
Moreover, examples of various types of relationships can be located, by referring to the relationship specified in Section \ref{411}.

\subsection{Class-aware Graph Sampling (CaGS)}\label{42}
In the CaGS module, CSRMS acquires a batch-level sub-graph from the dataset-level relation graph, based on sampling strategies. Moreover, curriculum construction is conducted based on easy-hard estimation to further enhance the representation learning.

\subsubsection{Positive Sampling from Dominant Patterns.}\label{421}
The relation $R_{ID}$ indicates samples of the same class may be spread out in the feature space. To help the model learn the representations of an image $I_i$ in a specific class $y_i$, CaGS uses dominant pattern sampler $\operatorname{S}_{dp}$ to select top-n positive samples that get the maximum distance from $I_i$ in the largest cluster $c_j$ dominated by class $y_i$:
\begin{equation}
    \Omega_{posi} = \operatorname{S}_{dp}(I_i,\varphi(I_i,I^{j}_{i}),n)
\end{equation}
where $\Omega_{posi}$ is the set of positive samples, and $\varphi(\cdot)$ calculates the distance between $I_i$ and the samples $I^{j}_{i}$ of class $i$ in cluster $c_j$.
Within a cluster, the number of images of a certain class can reflect the \emph{representativeness}, which is proportional to the number. So choosing positive samples from the biggest cluster enhances the chances of obtaining representative features of the class. Additionally, the distance condition improves the effectiveness of the selected positive samples in guiding.

 \subsubsection{Negative Sampling from Visually-similar Patterns.}
The inter-class relations $R_{IS}$ and $R_{MC}$ impact the classification mainly because of the visual similarity. Thus, CaGS uses a visually-similar pattern sampler $\operatorname{S}_{vp}$ to select top-m negative samples that get the minimum distance from $I_i$ in the nearest cluster $c_l$ dominated by class excluding $y_i$:
\begin{equation}
    \Omega_{nega} = \operatorname{S}_{vp}(I_i,\varphi(I_i,I^{l}_{i}),m)
\end{equation}
where $\Omega_{nega}$ is the set of negative samples, and $\varphi(\cdot)$ calculates the distance between $I_i$ and the samples $I^{l}_{i}$ of class $i$ in cluster $c_l$.

\subsubsection{Curriculum Learning for Batch Construction.}\label{423}
The previous section described how to collect positive and negative samples that can enhance representation learning through sample relations. However, introducing hard samples too early is not conducive to model learning. Therefore, in this section, we introduce batch construction based on curriculum learning.
Unlike traditional reinforcement learning-based methods \cite{c39,c45}, we quantify the \emph{representativeness} of clusters to estimate the difficulty of the samples. To be specific, for cluster $c_k$, we denote the number of images of class $y_j$ in the cluster by $N_{K}^{j}$ and denote the number of all images in the cluster by $N_K$. Thus, the \emph{representativeness} of cluster $c_k$ for class $y_j$ is denoted by $\frac{N_{K}^{j}}{N_K}$. And then, for an input image $I_i$ of class $y_j$, its level of difficulty is mapped to:
\begin{equation}
f(I_{i}) =
\begin{cases}
\Omega_{easy}  & \text { if } \frac{N_{K}^{j}}{N_K}>\rho_{1} \\
\Omega_{medium}, & \text { if } \forall \frac{N_{K}^{j}}{N_K}<\rho_{1}\\
\Omega_{hard}, & \text { if } \frac{N_{K}^{h}}{N_K}>\rho_{1} \text { and } j \neq h
\end{cases}
\end{equation}
where $\rho_1$ is a threshold that we set.

CSRMS takes a "decay method" to adjust the representation learning of samples. To be specific, we set different penalty coefficients for samples: $\lambda_e$ for easy, $\lambda_m$ for medium and $\lambda_h$ for hard. $\alpha_i$ and $\alpha_f$ are utilized to regulate the coefficients, defined by:
% α * y3 + (1 - α) * (α * y2 + (1 - α) * y1)
\begin{equation}
    \alpha_i*\lambda_e+(1-\alpha_i)*(\alpha_f*\lambda_m+(1-\alpha_f)*\lambda_h)=1
\end{equation}
Empirically, $\alpha_i$ and  $\alpha_f$ are initialized close to 1. When the loss converges to less than 0.01 and the average loss difference between two consecutive rounds of iteration is less than 0.0001, we start to decrease $\alpha_i$. Similarly, when the loss converges again, we start to decrease $\alpha_f$.

\subsection{Relational Graph-Guided Representation Learning (RGRL)} \label{431}
Under the guidance of the relational graph constructed in Section 4.1 and the curriculum constructed in Section 4.2, RGRL explicitly aggregates the representations of images of the same class and constrains the distance between representations of images in different classes to better alleviate the negative impact of visual noise. 

\subsubsection{Cluster-aware Representation Smoothing.}\label{431}
This module aims to perform batch-level and cluster-level representation smoothing by aggregating the information of representations, thereby alleviating the “intra-class diversity”.

\textbf{Graphical smoothing:} In order to aggregate the information of images of the same class to complete intra-class representation smoothing, CSRMS utilizes a graphical smoothing $\mathcal{G}(\cdot)$, using batch-level subgraph as knowledge to guide the information aggregation between input image $I_j$ and positive samples to generate enhanced representation. KNN algorithm is utilized to construct the symmetric adjacency $\hat{A}$ between the representation of $I_j$ and the representation of positive samples.
\begin{equation}
    F=\operatorname{concat}(\mathcal{M}_{v}(I_j), \mathcal{M}_{v}(\Omega_{posi}))
\end{equation}
\begin{equation}
    F_g=\mathcal{G}\left(\hat{A}, I_j, \Omega_{posi}\right)= \text{softmax}\left(\hat{A}\ \text{ReLU}\left(\hat{A} XW^{(1)}\right)W^{(0)}\right)
\end{equation}
where $\mathcal{M}_{v}(\cdot)$ denotes the visual encoder, $\hat{A}$ is symmetric adjacency, $W^{(0)} \in \mathbb{R}^{C \times H}$ denotes the input-to-hidden weight matrix for a hidden layer with $H$ feature maps and $W^{(1)} \in \mathbb{R}^{H \times F}$ denotes the hidden-to-output weight matrix.

\textbf{Cluster-level feature smoothing:}
In order to further aggregate representations of the same class at the cluster-level, CSRMS utilizes a cluster-level alignment to explicitly aggregate representations $F_g$ and cluster prototype $w_{cu}$ to generate aligned representations.
% The generation of $w_{cu}$ is based on the representativeness of the clusters. 
Specifically, for $I_i$ of class $y_j$, CSRMS aggregates the representations in the  largest cluster $c_j$ dominated by class 
$y_j$ to get the cluster prototype $w_{cu}^{j}$.
The alignment process is defined by:
\begin{equation}
    F_u=\alpha_u \odot F_{g} + \beta_u \odot w_{cu}
\end{equation}
where $\odot$ denotes the dot product and $\alpha_u$ and $\beta_u$ are the coefficients.

\subsubsection{Class-level Distribution Regularization.}
This module aims to complete the instance-level and class-level constraints by explicitly constructing dispersion loss, thereby alleviating the “inter-class similarity”.

\textbf{Class-level Representation Alignment:}
This module aims to complete representation smoothing by aggregating information at the class-level, thereby further alleviating the “intra-class diversity” of images.
For $I_i$ of class $y_j$, we aggregate the representations in all clusters dominated by class $y_j$ to get the class-level prototype $p_{ca}$.
To be specific, in Section \ref{423}, we quantify the \emph{representativeness} of cluster $c_k$ for class $y_j$: $\frac{N_{K}^{j}}{N_K}$. If $\frac{N_{K}^{j}}{N_K} >\rho_2 $, we identify the cluster $c_k$ is dominated by  class $y_j$.
The alignment process is defined by:
\begin{equation}
    F_a=\alpha_a \odot F_{u} + \beta_a \odot p_{ca}
\end{equation}
where $\odot$ denotes the dot product and $\alpha_a$ and $\beta_a$ are the coefficients.

After that, CSRMS learns a classifier for class prediction and uses the cross-entropy loss as the supervised loss defined by:
\begin{equation}
    \mathcal{L}_{ce}=\frac{1}{N} \sum_{i} \mathcal{L}_{i}=-\frac{1}{N} \sum_{i} \sum_{c=1}^{C} y_{i c} \log \left(P_{i c}\right)
\end{equation}
where $P$ denotes the prediction of CSRMS, $y$ denotes the labels of images and $C$ denotes the number of classes.

\textbf{Negative sampling constraint:} CSRMS constructs loss $\mathcal{L}_{nega}$ between the representation of the input image $I_{j}$ and the representations of negative samples $I_{nega}$, explicitly constraining the distance between representations, defined by:
\begin{equation}
    \mathcal{L}_{nega }=\sum_{i=1}^{N}-\log \left(\mu_{i} \frac{\theta}{\sum_{q=1}^{m}\left\|\mathcal{M}_{v}\left(I_{i}\right)-\mathcal{M}_{v}\left(\Omega_{nega}^{q}\right)\right\|_{2}+\theta}\right)
\end{equation}
where $\mathcal{M}_{v}(\cdot)$ denotes the visual encoder, $\theta$ denotes a fixed parameter, $N$ and $m$ denotes the number of images in the dataset and the number of negative samples.

\textbf{Inter-class constraint:} On this basis, CSRMS constructs dispersion loss between smoothed and aligned representations $F_a$ of different categories, and explicitly constrains the distance of inter-class representations.
\begin{equation}
    F_{v}^{j}=\sigma_ {v}\left(\sum_{1}^{\mathrm{m}} F_{a}^{j}\right)
\end{equation}
\begin{equation}
   \mathcal{L}_{inter}=\sum_{i, j}\{i=j\}\|\textbf{0}\|+\{i \neq j\}-\log \left(\mu \frac{\theta}{\left.\left\| F_{v}^{i}-F_{v}^{j}\right\|_{2}+\theta\right)}\right)
\end{equation}
where $\sigma_v$ denotes the aggregation process, $F_{a}^{j}$ denotes the representation after the class-level alignment of class $j$ and $\theta$ denotes a fixed parameter.

\begin{table}[t]
  \caption{Statistics of the datasets used in the experiments.}
  \vspace{-0.4cm}
  \label{tdataset}
  \begin{tabular}{ccccc}
    \toprule
    Datasets& \#Classes& \#Image Size & \#Training  & \#Testing \\
    \midrule
    CIFAR10 & 10 &	32*32&	 50,000&	10,000\\
    CIFAR100& 100&  32*32&	 50,000&	10,000\\
    Vireo172& 172&	224*224& 66,114     &  33,072    \\
    NUS-WIDE&  81&	224*224& 121,962	&81,636\\
  \bottomrule
\end{tabular}
 \vspace{-0.5cm}
\end{table}

\begin{table*}[t]
\renewcommand\arraystretch{1.0}
  \caption{Performance comparison of algorithms. Metrics are Top-1/Top-5 Accuracy (Acc), Precision (P), and Recall (R).}
  \label{tcomparison}
  \begin{tabular}{c|c|c|c|c|c|c|c|c|c|c}
    \hline
    \multirow{2}*{Algorithm} & \multicolumn{2}{|c}{CIFAR10}  &  \multicolumn{2}{|c}{CIFAR100} & \multicolumn{2}{|c}{Vireo172}& \multicolumn{4}{|c}{NUS-WIDE}\\
    \cline{2-11}
    ~ &  ACC@1 & ACC@5  &  ACC@1 &  ACC@5 &  ACC@1 &  ACC@5& P@1 &P@5 &R@1 &R@5  \\
    \hline
    LeNet5  & 72.77 & 97.20 & 40.54 & 80.73       & 20.33 &  42.77& 41.86 & 24.36 & 21.45 & 53.67 \\
    ResNet18 & 90.70 & 99.60 & 70.12 & 90.80       &76.31  &93.26& 73.71 & 37.19 & 40.12 & 79.94 \\
    ResNet50 & 92.03 & 99.79 & 71.97 & 91.44       &77.64  &93.55& 73.78 & 37.29 & 40.30 & 80.13 \\
    RE (AAAI'2020)        & 93.21 & 99.82 & 72.16 & 91.98    &78.49  &93.75& 74.18 & 37.46 & 40.66 & 80.36 \\
    DLSA (ECCV'2022)     & 93.37  & 99.74  & 72.31  & 92.40  &78.55  &93.77& 74.21 & 37.43 & 40.73 & 80.21 \\
    ACmix (CVPR'2022)     & 93.43 & 99.85 & 72.27 & 92.24    &78.85  &93.11& 74.33 & 37.53 & 40.92 & 80.27 \\
    Resizer (ICCV'2021)   & 93.49 & 99.83 & 73.01 & 92.16    &79.37  &93.98& 74.49 & 37.25 & 41.25 & 80.33 \\
    BatchFormer (CVPR'2022) & 93.54 & 99.84 & 73.13 & 92.76    &79.96  &94.20& 74.52 & 37.68 & 41.63 & 80.72 \\
    CUDA (ICLR'2023)      & 93.58 & 99.86 & 73.66 & 92.68    &81.13  &94.57& 74.60 & 38.26 & 42.02 & 80.86 \\
    ViT (ICLR'2021)    & 98.68 & 99.99 & 81.70 & 96.02       &85.92  &96.47& 79.75 & 39.86 & 44.64 & 86.10 \\
    \hline
    CSRMS(LeNet5)  & 74.88 & 98.01 & 44.77 & 82.15 &22.47	&45.47& 44.26 & 26.87 & 26.37 & 59.33 \\
    CSRMS(ResNet18)& 94.62 & 99.99 & 74.57 & 94.47 &82.60	&95.92& 75.12 & 39.42 & 42.25 & 81.17 \\
    CSRMS(ResNet50)& 95.48 & 99.99 & 76.13 & 95.51 &84.72	&96.87& 75.33 & 40.12 & 42.41 & 81.20 \\
    \textbf{CSRMS(ViT)}     & \textbf{99.44} & \textbf{99.99} & \textbf{84.93} & \textbf{98.08} & \textbf{88.99} & \textbf{98.85}& \textbf{80.68} & \textbf{41.33} & \textbf{46.02} & \textbf{87.29} \\
    \hline
  \end{tabular}
  \vspace{-0.3cm}
\end{table*}

\section{EXPERIMENTS}
\subsection{Experiment Settings}
\subsubsection{Datasets.}
In order to verify the effectiveness of CSRMS, we study our models based on publicly available datasets: CIFAR10, CIFAR100, Vireo172, and NUS-WIDE, and the statistics are shown in Table 1. In detail, CIFAR10 and CIFAR100 are colour image datasets that are closer to a universal object, both containing 60000 image samples.  Vireo172 comprises a collection of food images that feature various Chinese dishes, containing 110,241 image samples, we refer to the original paper \cite{c30} and divide the training/testing set. NUS-WIDE is a multi-label classification dataset, originally containing 269,648 image samples, we refer to the original paper \cite{c31} and related works \cite{c32,c33} to divide the training/testing set and remove the sample without label or tag.

\subsubsection{Evaluation Protocol.}
For the single-label datasets CIFAR10, CIFAR100 and Vireo172, we followed conventional measures of Top-1 and -5 accuracies to evaluate the classification performance. While for the NUS-WIDE multi-label dataset, we followed the original setups \cite{c31} to use Top-1 and -5 precision and recall.

\subsubsection{Implementation Details.}\label{513}
To verify the applicability of CSRMS, we investigate the performance of CSRMS on four visual backbones LeNet5 \cite{c34}, ResNet18 \cite{c7}, ResNet50 \cite{c7} and ViT \cite{c8} denoted as CSRMS(LeNet5), CSRMS(ResNet18), CSRMS(ResNet50) and CSRMS(ViT). The batch-size is fixed at 32. During training, we choose to use the SGD optimizer and the learning rate is selected from 5e-3 to 1e-1. The decay rate of the learning rate is selected from 0.1 and 0.5, and the decay interval is 20 epochs. The distance mentioned in 4.2 can be expressed using Euler distance or cosine similarity. The number of positive samples $n$ and negative samples $m$ are chosen from [5,10,20].
Regarding the threshold $\rho_1$ in Curriculum Construction, we conducted multiple experiments and choose it from 0.75 to 0.85 for the CIFAR100 and Vireo172 datasets, and choose it from 0.85 to 0.95 for the CIFAR10 and NUS-WIDE datasets. Moreover, for threshold $\rho_2$ mentioned in Class-level Representation Alignment, it is chosen from 0.5 to 0.55 for the CIFAR100 and Vireo172 datasets and chosen from 0.55 to 0.65 for the CIFAR10 and NUS-WIDE datasets based on experimental results.

\subsection{Performance Comparison}
This section reports the experimental performance of CSRMS and various baseline algorithms for image classification. The algorithms were evaluated using widely-used visual backbones, including LeNet \cite{c34}, ResNet-18 \cite{c7}, ResNet-50 \cite{c7} and ViT \cite{c8}, as well as state-of-the-art algorithms such as RE \cite{c37}, ACmix \cite{c9}, Resizer \cite{c38}, BatchFormer \cite{c14}, DLSA \cite{c40} and CUDA \cite{c39} combined with ResNet18. The hyperparameters for CSRMS and the baselines were carefully tuned to achieve the best performance. From the performance as reported in Table \ref{tcomparison}, we can observe the followings:
% This section reports the experimental performance of CSRMS and baseline algorithms for image classification in two categories: 1) the generally-used visual backbones including LeNet \cite{c34}, ResNet-18 \cite{c7}, ResNet-50 \cite{c7} and ViT \cite{c8}; 2) the state-of-the-art algorithms including RE \cite{c37}, ACmix \cite{c9}, Resizer \cite{c38}, BatchFormer \cite{c14}, DLSA \cite{c40} and CUDA \cite{c39}, all of which are combined with ResNet18. The hyperparameters of CSRMS and all the baselines were tuned to obtain the best performance by following Section \ref{513}.
% From the performance as reported in Table \ref{tcomparison}, we can observe the followings:
\vspace{-0.1cm}
\begin{itemize}[leftmargin=10pt]
    \item \textbf{The proposed CSRMS method consistently improves the generalization ability of visual backbones.} In different domains and datasets of different sizes, CSRMS can combine various types of backbone, including convolution-based models and Transform-based models, to improve classification performance.
    \item \textbf{CSRMS achieves more significant improvement in the case of high class complexity.} Most of the methods achieve good performance when facing the CIFAR-10 dataset with low classification complexity, while on CIFAR-100, VireoFood-172, and NUS-WIDE with the higher number of categories, CSRMS addresses the intra-class diversity and inter-class similarity problems, thus obtaining a 3\%-11\% improvement.
    \item \textbf{Visual backbones with more complex structures demonstrate significant advantages on larger datasets.} In the case of models with relatively basic architectures, such as LeNet5, there is a higher likelihood of overfitting when training on datasets with a higher number of categories such as CIFAR-100, or with a larger number of samples such as VireoFood-172 and NUS-WIDE. And the performance gap is more pronounced when comparing ResNet and ViT models.
    \item \textbf{Methods that introduce sample relationship or class-aware information can improve classification performance.} For example, compared with the advanced data augmentation method RE, the performances of CUDA that combine class-aware information have been improved for 2\%-4\% on larger datasets VireoFood-172 and NUS-WIDE; for relational modelling methods, BatchFormer and ACmix, which increases attention between samples, also make performance gains.
\end{itemize}

\vspace{-0.4cm}
\subsection{Ablation Study}
In this section, we further studied the working mechanisms of different modules of CSRMS, as shown in Table \ref{tablation}. The following findings could be observed:
\begin{itemize}[leftmargin=10pt]
\item \textbf{Positive sampling with graphical smoothing can alleviate intra-class diversity:} the accuracy obtained after adding DomPattern-based Intra-Class Sampling (CS(D)) and Graphical Smoothing (G) is always better than base, which verifies that graphical smoothing can alleviate effectiveness in terms of intra-class diversity.
\item \textbf{Negative sampling with explicit constraints can mitigate inter-class similarity:} adding "+CS(S)" can push away representations of different categories through explicit constraints to enhance representation learning so that achieves better classification accuracy. 
\item \textbf{Curriculum construction can help the model better learn complex features and knowledge:} "+CS(D)+G+CS(S)+CS(C)" can significantly improve the accuracy of both models. This is because curriculum construction enhances representation learning by making the model converge faster and avoiding overfitting and local optimal solutions.
\item \textbf{Cluster-level, class-level alignment and inter-class constraints can further alleviate intra-class differences and inter-class similarities: }the introduction of "A" and "IC" can further improve the accuracy of the two models. This is because “A” and “IC” can provide aggregation and constraints of more explicit schemes, thus enhancing representation learning.
\end{itemize}

\begin{table}[t]
% \resizebox{}{}{}
\footnotesize
\renewcommand\arraystretch{1.3}
\setlength{\tabcolsep}{1.2pt}
  \caption{Ablation study of CSRMS with ResNet18 and ViT backbone. CS(D):Positive Sampling from Dominant Patterns; CS(S): Negative Sampling from Visually-simi; CS(C): Curriculum construction strategy; G: Graphical smoothing module; A: Cluster-level and class-level alignment; IC: Class-level Distribution Regularization.}
  \vspace{-0.3cm}
  \begin{tabular}{c|c|c|c|c|c}
    \hline
    \multirow{2}*{Backbone}&\multirow{2}*{Models} & \multicolumn{2}{|c}{CIFAR100}  &  \multicolumn{2}{|c}{Vireo172} \\
    \cline{3-6}
    ~ & ~ &  ACC@1 & ACC@5  &  ACC@1 &  ACC@5   \\
    \hline
    \multirow{6}*{ResNet18} &Base	                   &70.12	&91.80	&77.31	&93.26\\
    \cline{2-6}            ~&+CS(D)+G	               &71.65	&93.76	&78.69	&93.82\\
    \cline{2-6}            ~&+CS(D)+G+CS(S)	           &72.95	&94.01	&80.62	&94.03\\
    \cline{2-6}            ~&+CS(D)+G+CS(S)+CS(C)	   &73.30	&94.28	&81.25	&94.96\\
    \cline{2-6}            ~&+CS(D)+G+CS(S)+CS(C)+A	   &73.62	&94.40	&81.79	&95.38\\
    \cline{2-6}            ~&+CS(D)+G+CS(S)+CS(C)+A+IC &74.57 & 94.47 &82.60	&95.92\\

    \hline
    \multirow{6}*{ViT}   &Base	                        &81.70 	&96.02	&85.92	&96.47\\
    \cline{2-6}         ~&+CS(D)+G	                    & 82.55	& 96.51	& 86.60	&97.32\\
    \cline{2-6}         ~&+CS(D)+G+CS(S)	            & 83.68	&97.04	&87.61	&98.52\\
    \cline{2-6}         ~&+CS(D)+G+CS(S)+CS(C)	        &84.05	&97.43	&88.10	&98.69\\
    \cline{2-6}         ~&+CS(D)+G+CS(S)+CS(C)+A	    &84.42	&97.85	&88.45	&98.77\\
    \cline{2-6}         ~&\textbf{+CS(D)+G+CS(S)+CS(C)+A+IC}     &\textbf{84.93} & \textbf{98.08} & \textbf{88.99} & \textbf{98.85}\\

    \hline
  \end{tabular}
  \label{tablation}
  \vspace{-0.4cm}
\end{table}

\vspace{-0.2cm}
\subsection{In-depth Analysis}
\subsubsection{Analysis of the Impact of Different Sampling Strategies on Classification Accuracy.}
As illustrated in Table \ref{tindepth1}, we have evaluated the influence of distinct sample sampling strategies on the classification outcomes. \textbf{Positive sampling from images with dominant patterns (DC-posi) can better alleviate intra-class diversity:} "OC-posi" involves positive sampling from clusters other than the input image's own cluster, surpassing random sampling in mitigating intra-class diversity and improving classification. "DC-posi," on the other hand, goes a step further by selecting positive samples from the most representative cluster, unifying sample representation learning, reducing intra-class differences, and boosting classification performance. \textbf{Negative sampling from images with visually similar patterns (SC-posi) can better alleviate inter-class similarity:} "OC-nega" employs negative sampling from different clusters for the input image. This strategy leads to an "attempt to make sufficiently dissimilar images even more dissimilar", which results in a slight decrease in performance compared to random sampling. On the other hand, "SC-nega" selects negative samples from the same cluster as the input image. This targeted approach effectively addresses inter-class similarity issues, outperforming random sampling and leading to better classification results.

\begin{table}[htbp]
\renewcommand\arraystretch{1.0}
\setlength{\tabcolsep}{2.5pt}
  \caption{Ablation of different sampling strategies.}
  \vspace{-0.4cm}
  \begin{tabular}{c|c|c|c|c|c}
    \hline
    \multirow{2}*{Backbone}&\multirow{2}*{Strategies} & \multicolumn{2}{|c}{CIFAR100}  &  \multicolumn{2}{|c}{Vireo172} \\
    \cline{3-6}
    ~ & ~ &  ACC@1 & ACC@5  &  ACC@1 &  ACC@5   \\
    \hline
    \multirow{7}*{ResNet18} & Base     & 70.12 & 91.80 & 77.31 & 93.26 \\
    \cline{2-6}            ~&  Random-posi & 70.82	&91.99	&77.69	&93.42\\
    % \cline{2-6}            ~&  SC-posi     & 70.45	&91.84	&77.44	&93.27\\
    \cline{2-6}            ~&  OC-posi     & 71.06	&92.03	&77.94	&93.58\\
    \cline{2-6}            ~&  DC-posi    & 71.65	&93.76	&78.69	&93.82\\
    \cline{2-6}            ~&  Random-nega      & 71.82	&93.85	&79.03	&93.67\\
    \cline{2-6}            ~&  OC-nega     & 71.77	&93.82	&78.82	&93.63\\
    \cline{2-6}            ~&  SC-nega    & 72.95	&94.01	&80.62	&94.03\\
    \hline
    \multirow{7}*{ViT} & Base         &81.70 	&96.02	&85.92	&96.47\\
    \cline{2-6}        ~&  Random-posi     & 81.99	&96.20	&86.17	&96.86\\
    % \cline{2-6}        ~&  SC-posi         & 81.82	&93.56	&86.01	&96.70\\
    \cline{2-6}        ~&  OC-posi         & 82.24	&96.36	&86.37	&96.99\\
    \cline{2-6}        ~&  DC-posi      & 82.55	& 96.51	& 86.60	&97.32\\
     \cline{2-6}           ~&  Random-nega    & 82.76	&96.62	&86.84	&97.79\\
    \cline{2-6}           ~&  OC-nega       & 82.73	&96.90	&86.80	&97.77\\
    \cline{2-6}           ~&  \textbf{SC-nega }      & \textbf{83.68}	&\textbf{97.04}	&\textbf{87.61}	&\textbf{98.52}\\
    \hline
  \end{tabular}
  \label{tindepth1}
  \vspace{-0.2cm}
\end{table}

\subsubsection{Analysis of the Impact of Different Smoothing Algorithms on Classification Accuracy.}
As illustrated in Table \ref{tindepth3}, we have evaluated the influence of distinct Smoothing algorithms on the classification outcomes.
% \begin{itemize}[leftmargin=10pt]
\textbf{Graphical smoothing can better aggregate the representation of the same class and better improve the classification effect than explicit constraint:} Utilizing graphical smoothing algorithms: GNN \cite{c42}, Tail-GNN \cite{c43} and GCN \cite{c44} to aggregate representations can achieve better classification accuracy than using explicit constraints such as js-divergence.
\textbf{Incorporating convolution when aggregating node information provides stronger expressive power and generalization ability:} GNN computes a weighted sum of adjacent nodes' features, while GCN uses convolutional operations for better representation fusion. This enables GCN to fuse relation information more effectively, resulting in better representation aggregation. Tail-GNN, though useful for imbalanced data, achieves similar performance to conventional GNN as it focuses on handling difficult samples rather than optimizing representation learning.
% \end{itemize}

\begin{table}[t]
\renewcommand\arraystretch{1.0}
\setlength{\tabcolsep}{2.5pt}
  \caption{Performance of CSRMS with smoothing algorithms.}
  \vspace{-0.4cm}
  \begin{tabular}{c|c|c|c|c|c}
    \hline
    \multirow{2}*{Backbone}&\multirow{2}*{Algorithms} & \multicolumn{2}{|c}{CIFAR100}  &  \multicolumn{2}{|c}{Vireo172} \\
    \cline{3-6}
    ~ & ~ &  ACC@1 & ACC@5  &  ACC@1 &  ACC@5   \\
    \hline
    \multirow{5}*{ResNet18} & Base           & 70.12	&91.80	&77.31	&93.26\\
    \cline{2-6}            ~&  JS-loss       & 70.88	&92.35	&78.24	&93.46\\
    \cline{2-6}            ~&  GNN           & 71.33	&92.87	&79.09	&94.12\\
    \cline{2-6}            ~&  Tail-GNN      & 71.38	&92.85	&79.13	&94.01\\
    \cline{2-6}            ~&  GCN            &71.65	&93.76	&78.69	&93.82\\

    \hline
    \multirow{5}*{ViT}     & Base        & 81.70 	&93.38	&85.92	&96.47\\
    \cline{2-6}            ~&  JS-loss   & 82.06	&93.51	&86.65	&96.76\\
    \cline{2-6}           ~&  GNN        & 82.25    &93.89	&86.88	&97.05\\
    \cline{2-6}           ~&  Tail-GNN   & 82.27    &93.88	&86.91	&97.05\\
    \cline{2-6}           ~&  \textbf{GCN} & \textbf{82.55}	& \textbf{96.51}& \textbf{87.69}	&\textbf{97.32}\\

    \hline
  \end{tabular}
  \label{tindepth3}
  \vspace{-0.2cm}
\end{table}

\subsubsection{Analysis of the Impact of Different Clustering Parameters on Classification Accuracy.}
\textbf{Based on our evaluation (Table \ref{tindepth4}), different clustering results minimally impact the final classification accuracy}. Although varying clustering outcomes can affect image recognition accuracy, the impact is limited because our clustering relies on visual commonality, and erroneous clusters constitute only a small fraction. These inaccuracies diminish gradually with increased training epoch iterations, as the network learns more precise feature representations and corrects erroneous clusters. Thus, for instances of poor clustering quality, excessive concern is unnecessary; instead, we should focus on enhancing the accuracy and generalization capacity of feature expression to achieve superior recognition results.
% We have evaluated the influence of different clustering results on the classification outcomes. According to table \ref{tindepth4}, we can conclude that \textbf{different clustering results have very little effect on the final classification accuracy.}
% Although different clustering results may affect the accuracy of image recognition, since our clustering is based on visual commonality, and these wrong clusters are only a small number, they have little effect on the final classification results. In addition, these erroneous clusters will gradually be weakened as the number of iterations of the training epoch increases, because, during the training process, the network can correct the erroneous clusters by learning more accurate feature representations. Therefore, for the case of poor clustering quality, we don't need to worry too much but should pay more attention to how to improve the accuracy and generalization ability of feature expression, so as to obtain better image recognition results.

\begin{table}[htbp]
\renewcommand\arraystretch{1.0}
\footnotesize
\setlength{\tabcolsep}{2.5pt}
  \caption{Image classification performance of CSRMS with different clustering parameters.}
\vspace{-0.3cm}
  \begin{tabular}{c|c|c|c|c|c}
    \hline
    \multirow{2}*{Backbone}&\multirowcell{2}{Vigilance\\ Parameter} & \multirowcell{2}{Number of\\ Clusters} & \multirowcell{2}{Number of\\Dominant Clusters} &  \multicolumn{2}{|c}{Vireo172} \\
    \cline{5-6}
    ~ & ~ & ~ & ~  &  ACC@1 &  ACC@5   \\
    \hline
    \multirow{5}*{ResNet18} & Base       & -	&-	&77.31	&93.26\\
    \cline{2-6}            ~&  0.5       & 503    &178	&82.56	&95.86\\
    \cline{2-6}            ~&  0.7       & 462	&207	&82.58	&95.90\\
    \cline{2-6}            ~&  0.85      & 424	    &220  &82.60	&95.92\\
    \cline{2-6}            ~&  0.95       & 406	    &261	&82.58	&95.90\\
    \hline
    \multirow{5}*{ViT}     & Base        & - 	&-	&85.92	&96.47\\
    \cline{2-6}            ~&  0.5       & 436	&185	&85.95	&98.84\\
    \cline{2-6}           ~&  0.7        & 398	&196	&88.95	&98.83\\
    \cline{2-6}           ~&  0.85 & 375	&203	&88.99	&98.85\\
    \cline{2-6}           ~&  0.95        & 369	&212	&88.95	&98.83\\
    \hline
  \end{tabular}
  \label{tindepth4}
  \vspace{-0.5cm}
\end{table}

\subsection{Case Study}
\subsubsection{Effect analysis of Relationship Modelling.}
\begin{figure}[t]
\centerline{\includegraphics[width=1.0\linewidth]{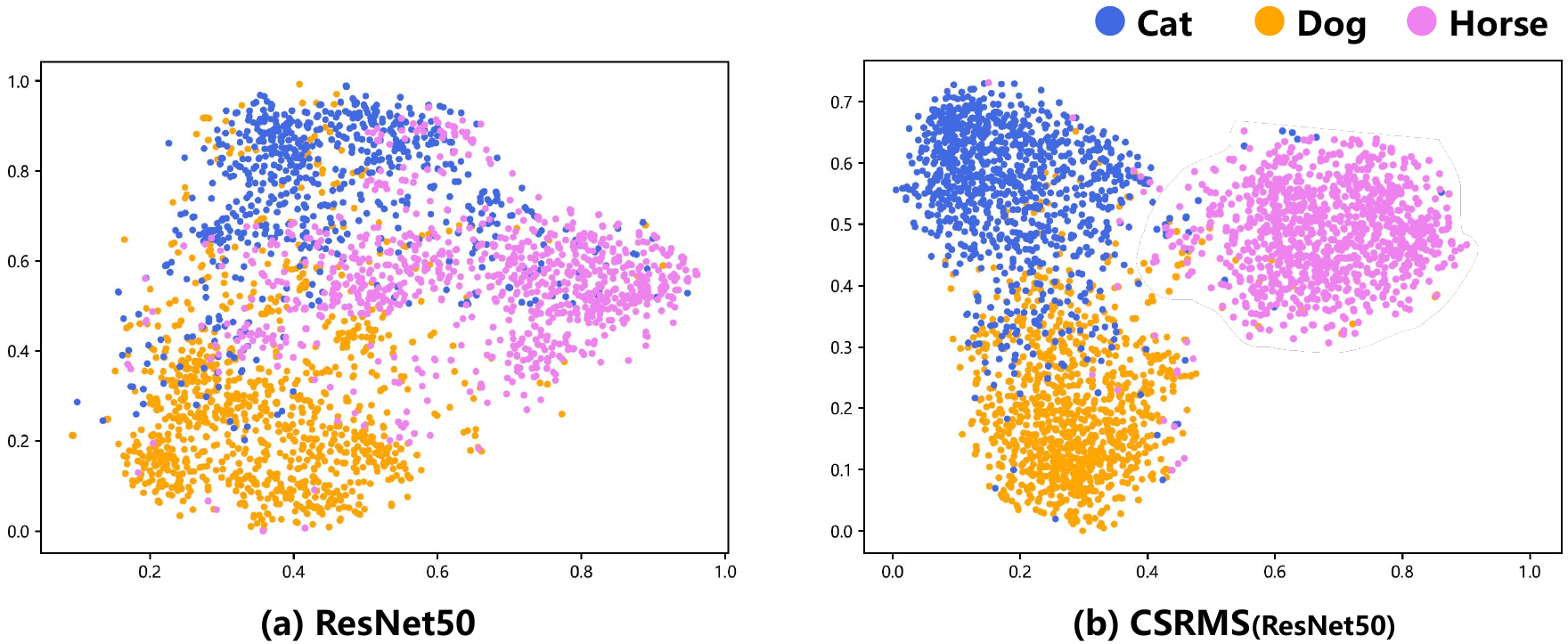}}
% \caption{2D t-SNE visualization of visual representations and error analysis. We choose the images with ground-truth labels "airplane" and "bird" from the test set of the CIFAR10 dataset and visualize the distribution for ResNet50 trained on CIFAR10 and CSRMS trained on CIFAR10.}
\vspace{-0.3cm}
\caption{Visualization of representations encoded by ResNet50 and CSRMS. To qualitatively verify the effect of introducing relational modelling on representation learning, we choose the images of three highly confusing categories: Cat, Dog and Horse, and compared the visualization results of representations encoded by ResNet50 and CSRMS.}
\label{fcase2}
\vspace{-0.7cm}
\end{figure}

This section evaluates the impact of relational modelling on representation learning. The confusion matrix analysis reveals that the cifar10 dataset's cat, dog, and horse categories experience the highest level of misclassification. To investigate this further, we select and visualize these images' representations encoded by ResNet50 and CSRMS using 2D-tsne, as presented in Figure \ref{fcase2}.
In Figure \ref{fcase2} (a), ResNet50 demonstrates relatively good discrimination among the three categories. However, the representation distribution across different categories is mixed, and representations of the same class are scattered widely. In contrast, Figure \ref{fcase2} (b) illustrates that CSRMS effectively addresses this issue by bringing representations of the same class closer together while pushing representations of different categories apart. This provides compelling evidence that introducing relationship modelling significantly enhances representation learning.
% This section evaluates the effect of introducing relational modelling on representation learning. By analyzing the confusion matrix, it is found that the images of three categories: cat, dog, and horse in the cifar10 dataset have the highest degree of confusion in classification. Therefore, we selected these images and visualized the representations encoded by ResNet50 and CSRMS using 2D-tsne, as shown in Figure \ref{fcase2}. As observed, in Figure \ref{fcase2} (a), although ResNet50 is able to distinguish images of the three categories relatively well, the distribution of representations belonging to different categories is heavily mixed, and the distribution of representations belonging to the same class is highly scattered. In contrast, in Figure \ref{fcase2} (b), CSRMS effectively mitigates this phenomenon by bringing the representation of the same class closer together and pushing away the representations of different categories. This fully validates the effectiveness of introducing relationship modelling in improving representation learning.
% \vspace{-0.5cm}

\vspace{-0.1cm}
\subsubsection{Error Analysis of Prediction.}
\begin{figure}[t]
\centerline{\includegraphics[width=1.0\linewidth]{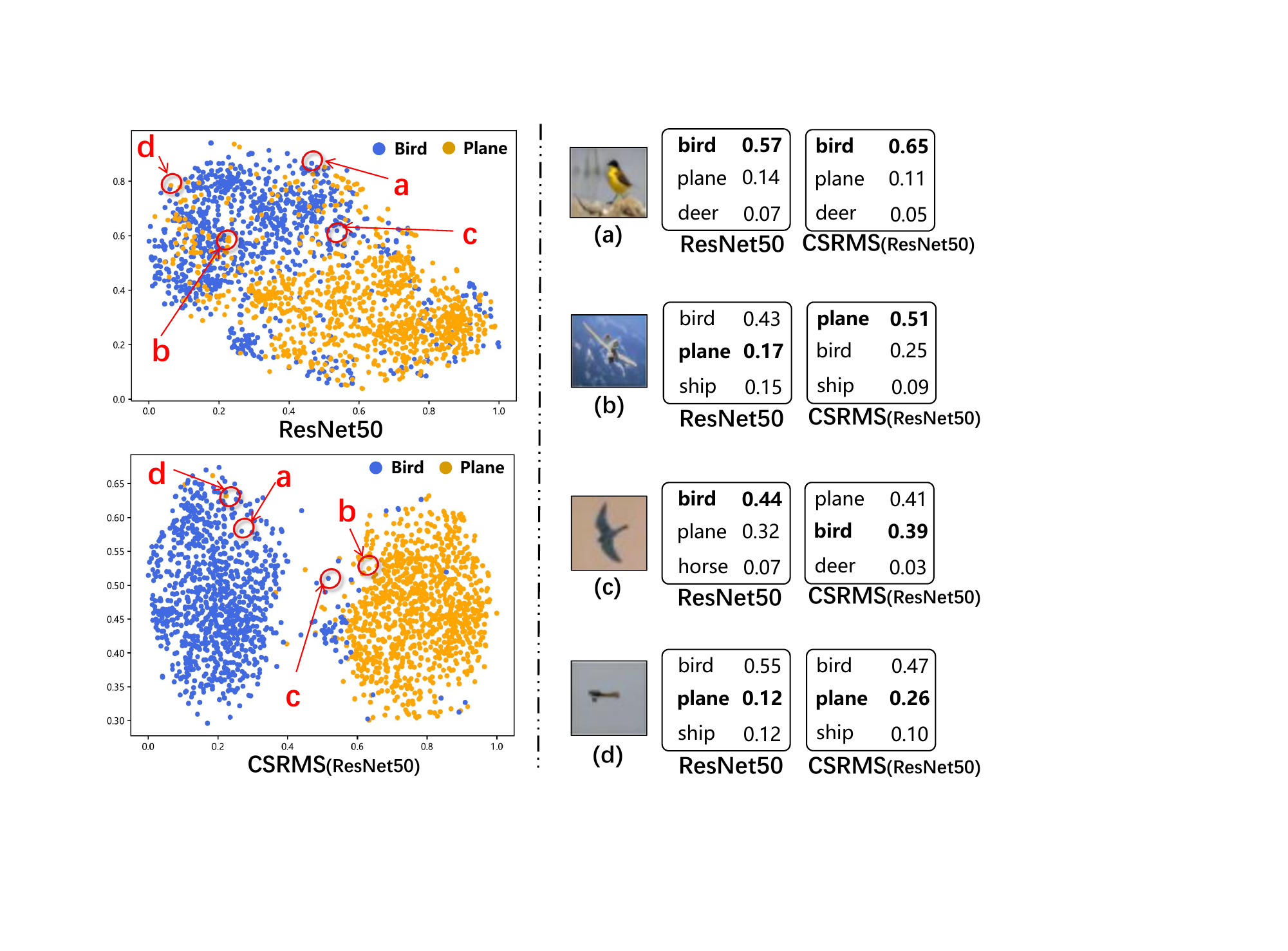}}
\vspace{-0.4cm}
% \caption{2D t-SNE visualization of visual representations and error analysis. We choose the images with ground-truth labels "airplane" and "bird" from the test set of the CIFAR10 dataset and visualize the distribution for ResNet50 trained on CIFAR10 and CSRMS trained on CIFAR10.}
\caption{Case Study on CSRMS in success and failure cases. To qualitatively demonstrate the efficacy of CSRMS in alleviating intra-class diversity and inter-class similarity, we select the images of two categories that are highly confused in the CIFAR10 dataset: “plane” and “bird”. (a) Both models achieve reasonable performance. (b) ResNet50 fails. (c) CSRMS performs worse. (d) both models perform badly.}
\vspace{-0.3cm}
\label{fcase}
\end{figure}

In this section, we present a 2D t-SNE visualization and error analysis to demonstrate the effectiveness of CSRMS in addressing intra-class diversity and inter-class similarity. Fig \ref{fcase} compares the visual representations generated by CSRMS and ResNet50 for four images, along with their predicted ingredients and confident scores.
In image (a), where the composition is clear, both models provide accurate predictions. However, CSRMS shows a higher confident score in predicting the "bird" position. Moreover, in the representation space, CSRMS places the representation of image (a) closer to the center of the "bird" representation set.
For image (b), where the components are unclear, ResNet50 tends to produce incorrect predictions. In contrast, CSRMS places the representation closer to the center of the "plane" representations set, leading to more accurate predictions.
In image (c), although both models produce incorrect predictions, CSRMS generates results highly similar to ResNet50. However, the representation space derived from ResNet50 exhibits an abnormally chaotic pattern, possibly occurring by chance.
For image (d), neither model accurately predicts the image due to its blurry components and small pixel range. Nevertheless, CSRMS places the representation closer to the "plane" representations set, resulting in a higher confident score.
These findings strongly support the efficacy of CSRMS in improving representation learning by mitigating intra-class diversity and inter-class similarity.

\section{CONCLUSION}
This paper proposes a novel approach CSRMS to alleviate the issue of intra-class visual diversity and inter-class similarity in representation learning by modelling a relational graph of the entire
dataset and performing class-aware smoothing and regularization operations. This approach learns the data distributions in the
feature space and extends the typical training batch construction process. A graph convolution network with knowledge-guided smoothing operations is utilized to ease the projection from different visual patterns to the same class. Experiments conducted on CIFAR10, CIFAR100, Vireo172, and NUS-WIDE datasets demonstrate the effectiveness of CSRMS in improving classification accuracy and verifying the effectiveness of structured knowledge modelling
for enhanced representation learning.

% Future work of this study may focus on incorporating rich multimodal information such as semantic information, which can help CSRMS to further optimize the sampling strategy, and can also help incorporate visual-semantic relationships to better aggregate representational information.
Future work of this study may focus on two directions. First, the model can integrate rich multimodal information, including semantic cues, to optimize the sampling strategy and foster a deeper understanding of visual-semantic relationships. Second, by incorporating self-supervised learning pre-training techniques, CSRMS can leverage unlabeled data to learn powerful representations, thereby improving data efficiency and generalization across tasks and domains. These avenues of exploration offer promising opportunities for the continued improvement of CSRMS.

\section{Acknowledgments}
This work is supported in part by the National Key R\&D Program of China (Grant no. 2021YFC3300203), the National Natural Science Foundation of China (Grant no. 62006141), the Oversea Innovation Team Project of the  "20 Regulations for New Universities" funding program of Jinan (Grant no. 2021GXRC073), the Excellent Youth Scholars Program of Shandong Province (Grant no. 2022HWYQ-048), and the TaiShan Scholars Program (Grant no. tsqn202211289).

% \begin{acks}
% \end{acks}

\bibliographystyle{ACM-Reference-Format}
\balance
\bibliography{sample-base}
% \bibliographystyle & \bibliography
% \bibliographystyle

% \bibliography
\end{document}